\documentclass[11pt,a4paper]{article}
\usepackage{authblk}
\usepackage[hyperref]{emnlp-ijcnlp-2019}
\usepackage{times}
\usepackage{latexsym}
\usepackage{booktabs}
\usepackage{bm} 
\usepackage{multicol}
\usepackage{multirow}
\usepackage{hyperref}
\usepackage{amsmath}
\usepackage{amssymb}
\usepackage{dsfont}
\usepackage{setspace}
\usepackage{tempora}
\usepackage{ifthen}

\usepackage{color, soul}
\usepackage[utf8]{inputenc}
\usepackage[russian,greek,english]{babel} 

\usepackage{transparent}

\usepackage{tikz}
\usetikzlibrary{positioning}
\usetikzlibrary{shapes}
\usetikzlibrary{decorations.pathreplacing}
\usetikzlibrary{arrows.meta}

\tikzset{font={\fontsize{9pt}{12}\selectfont}}
\tikzset{operator/.style={circle, draw, inner sep=0pt, minimum size=.8cm}}
\tikzset{mytip/.tip={>[length=12, width=16]}}

\definecolor{PaleBlue}{rgb}{0,.55,.9}
\definecolor{DarkBlue}{rgb}{0.02,.2,.6}
\definecolor{PaleGreen}{rgb}{0,.7,.25}
\definecolor{RedPink}{rgb}{.9,0,.2}
\definecolor{Pink}{rgb}{.85,.35,.7}
\definecolor{Purple}{rgb}{.6,0,.75}
\definecolor{DarkPurple}{rgb}{.21,0.17,.43}
\definecolor{Orange}{rgb}{.9,.3,.05}
\definecolor{GoldUL}{rgb}{1,.76,.32}

\colorlet{attentionColor}{Orange}
\colorlet{charEmbedColor}{RedPink}
\colorlet{predEmbedColor}{Pink}

\tikzstyle{embed}=[%
  draw,
  #1,
  anchor=north,
  minimum width=.3cm,
  minimum height=.8cm,
  inner sep=0pt,
  text=#1!65!black,
  thick,
  ]

\tikzstyle{comick_style}=[opacity=.6,line width=1pt, solid, #1, mark size=1.2pt, mark=*, mark options={#1, solid}]
\tikzstyle{mimick_style}=[opacity=.6,line width=1pt, #1, dashed, dash pattern=on 7pt off 1.5pt on 1pt off 1.5pt, mark size=1.2pt, mark=*, mark options={#1, solid}]

\newcommand{\att}[2]{%
\begin{tikzpicture}[baseline=(base), font={\fontsize{11pt}{12}\selectfont},
inner sep=0pt]%
\coordinate(base) at (0,-.5ex);
\node[anchor=mid, text=white](i) {\small #1};
\path (i.east) |- (0,0) coordinate[midway](right);
\path (i.west) |- (0,0) coordinate[midway](left);
\fill[red!#2] ([xshift=-.5*5.35 pt, yshift=1.5*5.35 pt]left) rectangle ([xshift=.5*5.35 pt, yshift=-1.5*5.35 pt]right);
\node at (i) {\small #1};
\end{tikzpicture}%
}

\usepackage{url}

\aclfinalcopy 

\title{Attending Form and Context to Generate Specialized Out-of-Vocabulary Words Representations}

\author[1]{\textbf{Nicolas Garneau}}
\author[1]{\textbf{Jean-Samuel Leboeuf}}
\author[2]{\textbf{Yuval Pinter}}
\author[1]{\textbf{Luc Lamontagne}}
\affil[1]{D\'epartement d'informatique et de g\'enie logiciel, Universit\'e Laval, Qu\'ebec}
\affil[2]{School of Interactive Computing, Georgia Institute of Technology}
\affil[ ]{\texttt{nicolas.garneau@ift.ulaval.ca, jean-samuel.leboeuf.1@ulaval.ca}}
\affil[ ]{\texttt{uvp@gatech.edu, luc.lamontagne@ift.ulaval.ca}}

\date{}

\begin{document}
\maketitle
\begin{abstract}
We propose a new contextual-compositional neural network layer\footnote{\url{https://github.com/ngarneau/contextual-mimick}} that handles out-of-vocabulary (OOV) words in natural language processing (NLP) tagging tasks.
This layer consists of a model that attends to both the character sequence and the context in which the OOV words appear.
We show that our model learns to generate task-specific \textit{and} sentence-dependent OOV word representations without the need for pre-training on an embedding table, unlike previous attempts.
We insert our layer in the state-of-the-art tagging model of \citet{plank2016multilingual} and thoroughly evaluate its contribution on 23 different languages on the task of jointly tagging part-of-speech and morphosyntactic attributes.
Our OOV handling method successfully improves performances of this model on every language but one to achieve a new state-of-the-art on the Universal Dependencies Dataset 1.4.
\end{abstract}


\section{Introduction}

The use of distributed word embeddings such as Word2Vec \cite{mikolov2013distributed}, Polyglot \cite{al2013polyglot}, and ELMo \cite{Peters2018DeepCW}, is widespread across NLP tasks such as part of speech tagging (POS) \cite{ling2015finding} and morphosyntactic attribute tagging (MORPH) \cite{cotterell2015morphological}.

Several of these embeddings databases have been trained on fixed size vocabulary corpus such as Wikipedia\footnote{\url{https://dumps.wikimedia.org}}.
When working on a downstream NLP task, systems relying on such embeddings become exposed to words that are not available in the embeddings database (out-of-vocabulary words, OOV).
This problem becomes more acute when considering deployment of a pre-trained model into production.
Improperly handling OOV words results in mapping words to embeddings with very low information, leading to reduced performance \cite{goldberg2017neural_chap8}.
The OOV problem is even more pronounced in low-resource languages, where we have limited access to a training dataset.
For example, in our setting, 35\% of the words of the Basque language miss an embedding.
Using our model results in a 5\% increase in OOV POS accuracy on this language.

The main contribution of this work is a new contextual-compositional model that learns alongside a sequence tagging neural network to generate useful task-specific and sentence-dependent word representations for OOV words.
It uses one BiLSTM that generates a hidden representation for each character and another one for the words of the context.
We apply an attention mechanism on each set of hidden representations that selects useful components to generate a final OOV word representation, which can then be used as a regular word embedding by the state-of-the-art tagging model.


We test this model on several languages and provide a thorough analysis of the results.
For a fair and meaningful comparison of our contribution, we employ a similar setup as \citet{pinter2017mimicking} and \citet{Zhao2018GeneralizingWE}, using the Polyglot embeddings \cite{al2013polyglot}, the same subset of the Universal Dependencies 1.4 corpus and the same tagging model \cite{plank2016multilingual}.
We evaluate our predicted word representations extrinsically on tasks of POS and MORPH tagging.

\section{Related work}

Handling out-of-vocabulary words has been an active field of research in the last few years.
A na{\"i}ve and simple way to handle OOV words is to assign it a random embedding, or to train a specific ``unknown'' embedding.
\citet{goldberg2017neural_chap8} emphasize that these methods have underestimated problems.


The Fasttext model \cite{bojanowski2016enriching} tackles this problem by modeling rare words as a bag of $n$-grams of characters.
While this representation may capture morphological phenomena such as suffixes and prefixes, it could not characterize properly the meaning of some words.
For example, the homograph \textit{bank} can mean a ``\textit{business that provides financial services}'' or a ``\textit{land along the side of a river or lake}'' depending on the context, which cannot be inferred by its characters. 

\newcite{pinter2017mimicking} proposed the Mimick model, which is a recurrent neural network trained on the character level using pre-trained embeddings such as Polyglot \cite{al2013polyglot} as the objective.
Once trained, the model can be used to predict word embeddings more useful than random embeddings for downstream NLP tasks.
A variant of Mimick called Bag of Substring \citep[BoSS]{Zhao2018GeneralizingWE} replaces the character sequence with a decomposition of the word into Fasttext-style ``bags of substrings'', whose representations are averaged to predict word embeddings.
However, this model also requires to be pre-trained on an embedding table before it can be used in a downstream task, and it does not take into account the context.

Recently, \newcite{Schick2018LearningSR} proposed an approach similar to Mimick and BoSS, but their model takes into account the context of a given word.
Given a huge body of text and a large embedding table, they gather every context in which a particular word appears to compute an \textit{aggregated} contextual representation for a word, making it as close as possible to its representation in the embedding table.
This approach also generate a non-contextual representation for a particular OOV.
The ELMo architecture proposed by \newcite{Peters2018DeepCW}, which leverages the morphology and the context, easily handles OOV words, but again, only given a large amount of training data.

In our setting, we specifically tackle the case where large-scale training data and resources are limited, such as foreign languages or specialized fields.
We emphasize that our model does not necessitate a pre-training step to learn to generate OOV representation, as in \cite{pinter2017mimicking} and \cite{Schick2018LearningSR},  where a mapping from a word to its pre-trained embedding is learned.
Our model instead learns to generate OOV word representations along with the target task's neural architecture and the corresponding training data.

\section{Generating and Interpreting Embeddings for OOV Words}

Our model aims to generate useful representations for OOV words as a substitute for embeddings, which depend on a specific NLP task.
Bearing this in mind, we first introduce our contextual-compositional OOV handling model, followed by the network that is used to perform the task, and finally the training details.

\subsection{Modeling OOV Words}

The model we propose to handle a word not present in the pre-trained vocabulary uses the characters of that particular word, as well as the context in which it appears.
For each target OOV word, a context window of $n$ words is first mapped to their Polyglot embeddings of dimension $64$.
If another OOV word appears in the context, a random embedding is assigned to it.
We assume that the impact of such words is negligible most of the time for a large enough context window, so we take up to a total of $n \leq 40$ words surrounding the target OOV word.

A Bidirectional Long Short Term Memory module \cite[BiLSTMs]{hochreiter1997long} is applied on the context and generates $n$ hidden states $\mathbf{h}$ of dimension $256$ (each direction has a hidden state of $128$).
The OOV word that is the subject of the prediction is simply skipped by the BiLSTM.
The attention mechanism, which is essentially a linear layer, is applied and outputs a score vector $\mathbf{s}$ of $n$ components.
These scores are then transformed into a distribution $\bm{\alpha} = \text{softmax}(\mathbf{s})$.
A vector representation of the context is computed by taking the weighted average of the hidden states, as $\mathbf{c} = \sum_{i=1}^{n} \alpha_i \mathbf{h}_i$. This procedure is depicted in the top left part of Figure~\ref{fig:oov_handling_net}.


A similar procedure is applied to the characters of the target OOV word, which yields a vector representation $\mathbf{w}$, as shown in the right part of Figure~\ref{fig:oov_handling_net}.
Characters embeddings are set to be of $20$ components.
We then concatenate $\mathbf{w}$ and $\mathbf{c}$ to form a vector representation of $512$ components which is fed to a two-layer linear network separated by a \texttt{tanh} activation function to obtain the final predicted word representation $\mathbf{p}$.
The first layer has shape $512\times64$, then the second has shape $64\times64$.
This vector representation $\mathbf{p}$ is then used in the downstream task's neural network as an ``embedding'' for the OOV word.
The bottom part of Figure~\ref{fig:oov_handling_net} illustrates the process.

\tikzset{operator/.style={circle, draw, inner sep=0pt, minimum size=.6cm}}
\tikzstyle{embed}=[%
    draw,
	#1,
	thick,
	anchor=north,
	minimum width=.3cm,
	minimum height=.4cm,
	inner sep=0pt,
	text=#1!65!black
	]
\tikzstyle{lstm}=[%
	draw,
	minimum width=.3cm,
	minimum height=.15cm,
	inner sep=0pt,
	anchor=north,
	thick,
]
\colorlet{colorOOV}{RedPink}
\colorlet{colorWordEmbeds}{Purple}
\colorlet{colorCharEmbeds}{DarkBlue}
\colorlet{colorAttention}{Orange}
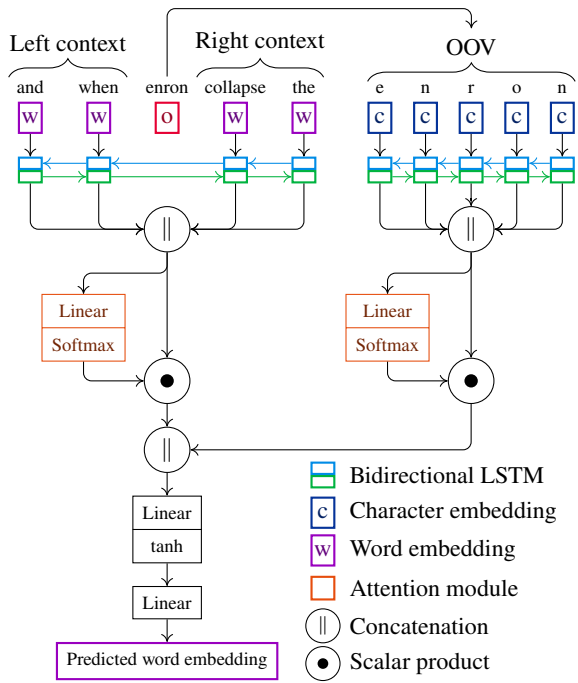
\begin{figure}[h]
\begin{tikzpicture}[%
	mytip/.tip={>[length=3, width=4.5]},
	scale=1,
	lstm right/.style={lstm, PaleGreen},
	lstm left/.style={lstm, PaleBlue},
	]
    
    	\def\embedWidth{.3}
    	\def\spaceBetweenChars{.3}
    	\def\spaceBetweenWords{.6}
    	\def\spaceBetweenContextAndOOV{4.0}
    
    	\def\spaceBetweenTextAndEmbed{.2}
    	\def\spaceBetweenEmbedAndLSTM{.3}
    	\def\spaceBetweenLSTMAndCat{.3}
    	\def\spaceBetweenCatAndFCSM{.55}
    	\def\xshiftFCSM{-1.1}
    	\def\spaceBetweenFCSMAndSP{.255}
    	\def\spaceBetweenOps{.3}
    
    	\def\roundedCorners{3pt}
    
    	\def\OOV{e,n,r,o,n}
    	\def\nChars{5}
    
    	\def\context{and, when, enron, collapse, the}
    	\def\nWords{5}
    	\def\posOOV{3}

	\begin{scope}[xshift=\spaceBetweenContextAndOOV cm]
	\foreach \letter [count=\i] in \OOV{
		\pgfmathsetmacro\pos{(\i-(\nChars+1)/2)*(\embedWidth+\spaceBetweenChars)}
		\node[anchor=mid, inner sep=0pt](w \i) at (\pos,0) {\scriptsize \letter};
		\node[embed=colorCharEmbeds](w embed \i) at (\pos,-\spaceBetweenTextAndEmbed) {\footnotesize c};
		\node[lstm left](w lstm left \i) at ([yshift=-\spaceBetweenEmbedAndLSTM cm]w embed \i.south) {};
		\node[lstm right](w lstm right \i) at (w lstm left \i.south) {};
		\draw[-mytip] (w embed \i) -- (w lstm left \i);
	}

	\draw[decorate,decoration={brace,amplitude=7pt}]
		([yshift=1pt,xshift=-.1cm]w 1.north west) -- node[midway, yshift=13pt](w){OOV} ([yshift=1pt,xshift=.1cm]w \nChars.north east);

	\foreach \letter [count=\i] in \OOV{
		\pgfmathsetmacro\iminusone{int(\i-1)}
		\pgfmathsetmacro\iplusone{int(\i+1)}

		\ifthenelse{\i=\nChars}{}{
			\draw[-mytip, PaleGreen] (w lstm right \i) -- (w lstm right \iplusone);
		}
		\ifthenelse{\i=1}{}{
			\draw[-mytip, PaleBlue] (w lstm left \i) -- (w lstm left \iminusone);
		}
	};

	\path ([yshift=-\spaceBetweenLSTMAndCat cm]w lstm right 1.south) -| (0,0) node[midway, operator, anchor=north](w cat) {$\|$};
	\foreach \letter [count=\i] in \OOV{
		\draw[-mytip, rounded corners=\roundedCorners] (w lstm right \i.south) |- (w cat);
	}

	\draw[-mytip, rounded corners=\roundedCorners] (w cat.south) |- +(\xshiftFCSM/2,-\spaceBetweenCatAndFCSM/2) -| +(\xshiftFCSM,-\spaceBetweenCatAndFCSM) coordinate(tmp FCSM);
	\node[draw=colorAttention, text=colorAttention!50!black, anchor=north, rectangle split, rectangle split parts=2](w FCSM) at (tmp FCSM){\scriptsize Linear \nodepart{second} \scriptsize Softmax};

	\node[operator](w SP) at ([xshift=-\xshiftFCSM cm, yshift=-\spaceBetweenFCSMAndSP cm]w FCSM.south) {};
	\fill (w SP) circle (2.5pt);
	\draw[-mytip, rounded corners=\roundedCorners] (w FCSM.south) |- (w SP);
	\draw[-mytip] (w cat) -- (w SP);
	\end{scope}

	\begin{scope}
	\pgfmathsetmacro\posOOVminusone{int(\posOOV-1)}
	\pgfmathsetmacro\posOOVplusone{int(\posOOV+1)}

	\foreach \word [count=\i] in \context{
		\pgfmathsetmacro\pos{(\i-(\nWords+1)/2)*(\embedWidth+\spaceBetweenWords)}
		\node[anchor=mid, inner sep=0pt](c \i) at (\pos,0) {\scriptsize \word};
		\ifthenelse{\i=\posOOV}{
			\node[embed=colorOOV](c embed \i) at (\pos,-\spaceBetweenTextAndEmbed) {\footnotesize o};
		}{
			\node[embed=colorWordEmbeds](c embed \i) at (\pos,-\spaceBetweenTextAndEmbed) {\footnotesize w};
			\node[lstm left](c lstm left \i) at ([yshift=-\spaceBetweenEmbedAndLSTM cm]c embed \i.south) {};
			\node[lstm right](c lstm right \i) at (c lstm left \i.south) {};
			\draw[-mytip] (c embed \i) -- (c lstm left \i);
		}
	}

	\draw[decorate,decoration={brace,amplitude=7pt}]
		([yshift=1pt,xshift=-.1cm]c 1.north west) -- node[midway, yshift=13pt]{Left context} ([yshift=1pt,xshift=.1cm]c \posOOVminusone.north east);
	\draw[decorate,decoration={brace,amplitude=7pt}]
		([yshift=1pt,xshift=-.1cm]c \posOOVplusone.north west) -- node[midway, yshift=13pt]{Right context} ([yshift=1pt,xshift=.1cm]c \nWords.north east);

	\draw[-mytip, rounded corners=\roundedCorners] ([yshift=.1cm]c \posOOV.north) |- +(2,.9) -| (w.north);

	\pgfmathsetmacro\nWordsminusone{int(\nWords-1)}
	\foreach \i in {1,...,\nWordsminusone}{
		\pgfmathsetmacro\next{int(\i+1)}
		\ifthenelse{\i=\posOOVminusone}{
			\pgfmathsetmacro\next{\posOOVplusone}
		}{}

		\ifthenelse{\i=\posOOV}{
		}{
			\draw[-mytip, PaleGreen] (c lstm right \i) -- (c lstm right \next);
		}
	}
	\foreach \i in {2,...,\nWords}{
		\pgfmathsetmacro\next{int(\i-1)}
		\ifthenelse{\i=\posOOVplusone}{
			\pgfmathsetmacro\next{\posOOVminusone}
		}{}

		\ifthenelse{\i=\posOOV}{
		}{
			\draw[-mytip, PaleBlue] (c lstm left \i) -- (c lstm left \next);
		}
	}

	\path ([yshift=-\spaceBetweenLSTMAndCat cm]c lstm right 1.south) -| (0,0) node[midway, operator, anchor=north](c cat) {$\|$};
	\foreach \word [count=\i] in \context{
		\ifthenelse{\i=\posOOV}{}{
			\draw[-mytip, rounded corners=\roundedCorners] (c lstm right \i.south) |- (c cat);
		}
	}

	\draw[-mytip, rounded corners=\roundedCorners] (c cat.south) |- +(\xshiftFCSM/2,-\spaceBetweenCatAndFCSM/2) -| +(\xshiftFCSM,-\spaceBetweenCatAndFCSM) coordinate(tmp FCSM);
	\node[draw=colorAttention, text=colorAttention!50!black, anchor=north, anchor=north, rectangle split, rectangle split parts=2](c FCSM) at (tmp FCSM){\scriptsize Linear \nodepart{second} \scriptsize Softmax};

	\node[operator](c SP) at ([xshift=-\xshiftFCSM cm, yshift=-\spaceBetweenFCSMAndSP cm]c FCSM.south) {};
	\fill (c SP) circle (2.5pt);
	\draw[-mytip, rounded corners=\roundedCorners] (c FCSM.south) |- (c SP);
	\draw[-mytip] (c cat) -- (c SP);

	\draw[-mytip] (c SP.south) -- +(0,-\spaceBetweenOps) node[operator, anchor=north] (cat wc) {$\|$};
	\draw[-mytip, rounded corners=\roundedCorners] (w SP) |- (cat wc);

	\draw[-mytip] (cat wc.south) -- ++(0,-\spaceBetweenOps) node[draw, rectangle split, rectangle split parts=2, anchor=north](FCtanh1) {\scriptsize Linear \nodepart{second} \scriptsize tanh};
	\draw[-mytip] (FCtanh1.south) -- ++(0,-\spaceBetweenOps) node[draw, anchor=north](FC) {\scriptsize Linear};
	\draw[-mytip] (FC.south) -- ++(0,-\spaceBetweenOps) node[draw, anchor=north, Purple, text=black, thick](output) {\scriptsize Predicted word embedding};
	\end{scope}

	\begin{scope}[xshift=2.05cm, yshift=-5.15cm]
	\def\legspace{-.5}
	\def\iconspace{.22}
	\node[draw, lstm left, anchor=south] (leg lstm left) at (0,0) {};
	\node[draw, lstm right, anchor=north] (leg lstm right) at (leg lstm left.south) {};
	\node[anchor=west] at (\iconspace,0) {Bidirectional LSTM};

	\node[embed=colorCharEmbeds, minimum height=.4cm, anchor=center](leg char embed) at (0,\legspace) {\footnotesize c};
	\node[anchor=west] at (\iconspace,\legspace) {Character embedding};

	\node[embed=colorWordEmbeds, minimum height=.4cm, anchor=center](leg word embed) at (0,2*\legspace) {\footnotesize w};
	\node[anchor=west] at (\iconspace,2*\legspace) {Word embedding};
	
	\node[embed=colorAttention, minimum height=.3cm, anchor=center](leg word embed) at (0,3*\legspace) {};
	\node[anchor=west] at (\iconspace,3*\legspace) {Attention module};

	\node[operator, minimum size=.45cm](leg cat) at (0,4*\legspace) {\scriptsize$\|$};
	\node[anchor=west] at (\iconspace,4*\legspace) {Concatenation};

	\node[operator, minimum size=.45cm](leg SP) at (0,5*\legspace) {};
	\node[anchor=west, align=left] at (\iconspace,5*\legspace) {Scalar product};
	\fill (leg SP) circle (2pt);
	\end{scope}

\end{tikzpicture}

\caption{Our contextual-compositional OOV handling neural network.}
\label{fig:oov_handling_net}
\end{figure}

\begin{table*}[h!]
\centering
\small
\def\cols{\hspace{3.5pt}}
\begin{tabular}{lc c@{\cols}c@{\cols}c@{\cols}c c@{\cols}c@{\cols}c c@{\cols}c@{\cols}c@{\cols}c c@{\cols}c@{\cols}c}
\toprule
\multirow{3}{*}[-4pt]{Language} & \multirow{3}{*}[-4pt]{OOV\%} & \multicolumn{7}{c}{POS} & \multicolumn{7}{c}{MORPH}\\
 & & \multicolumn{4}{c}{All} & \multicolumn{3}{c}{OOV} & \multicolumn{4}{c}{All} & \multicolumn{3}{c}{OOV}\\
\cmidrule(lr){3-6}
\cmidrule(rl){7-9}
\cmidrule(rl){10-13}
\cmidrule(l){14-16}
 & & Rand. & BoSS & Mim. & Ours & Rand. & Mim. & Ours & Rand. & BoSS  & Mim. & Ours & Rand. & Mim. & Ours\\
\cmidrule(lr){3-9}
\cmidrule(r){1-1}
\cmidrule(lr){2-2}
\cmidrule(l){10-16}
Kazakh		& 21\%	& 79.0	& 75.8	& 82.1	& \textbf{85.1}	& 56.8	& \textbf{80.2}	& 77.8	& 60.8	& 24.0	& 64.5	& \textbf{66.1}	& 35.6	& 47.5	& \textbf{54.2}\\
Tamil		& 16\%	& 82.9	& 77.4	& 86.7	& \textbf{87.7}	& 62.1	& 82.2	& \textbf{86.1}	& 87.9	& 76.2	& 90.1	& \textbf{91.9}	& 70.3	& 80.7	& \textbf{87.4}\\
Latvian		& 11\%	& 89.3	& 87.2	& 90.1	& \textbf{91.6}	& 61.4	& 71.7	& \textbf{79.9}	& 83.4	& 67.6	& 84.7	& \textbf{87.4}	& 61.7	& 67.8	& \textbf{79.9}\\
Vietnamese	& 22\%	& 88.2	& 84.6	& 88.9	& \textbf{90.0}	& 75.7	& 77.9	& \textbf{80.6}	& -		& -		& -		& -				& -		& -		& -\\
Hungarian	& 12\%	& 93.7	& 92.2	& 94.4	& \textbf{95.0}	& 77.6	& 86.1	& \textbf{91.0}	& 90.7	& 83.6	& 89.3	& \textbf{90.9}	& 79.7	& 85.2	& \textbf{90.5}\\
Turkish		& 15\%	& 93.4	& 89.0	& 94.2	& \textbf{95.2}	& 80.6	& 88.3	& \textbf{90.8}	& 94.1	& 82.6	& 94.4	& \textbf{95.3}	& 89.8	& 92.0	& \textbf{94.6}\\
Greek		& 18\%	& 97.7	& 96.5	& 98.1	& \textbf{98.5}	& 85.2	& 90.2	& \textbf{92.4}	& 96.5	& 93.4	& 97.0	& \textbf{97.4}	& 83.6	& 92.0	& \textbf{93.7}\\
Bulgarian	& 9\%	& 97.9	& 97.1	& 98.2	& \textbf{98.6}	& 89.2	& 92.9	& \textbf{95.1}	& 97.3	& 91.5	& 97.6	& \textbf{98.1}	& 91.5	& 94.2	& \textbf{95.6}\\
Swedish		& 9\%	& 96.3	& 94.5	& 97.1	& \textbf{97.6}	& 87.8	& 94.1	& \textbf{96.5}	& 95.8	& 93.0	& 96.8	& \textbf{97.4}	& 86.7	& 92.4	& \textbf{95.2}\\
Basque		& 35\%	& 93.8	& 91.3	& 94.6	& \textbf{95.8}	& 74.3	& 82.7	& \textbf{87.5}	& 92.6	& 82.0	& 93.5	& \textbf{95.1}	& 79.5	& 85.3	& \textbf{88.8}\\
Russian		& 17\%	& 95.7	& 94.8	& 96.6	& \textbf{97.2}	& 86.2	& 91.3	& \textbf{93.0}	& 92.7	& 91.5	& 93.5	& \textbf{94.9}	& 79.9	& 86.1	& \textbf{88.3}\\
Danish		& 8\%	& 95.8	& 94.7	& 96.3	& \textbf{96.5}	& 77.6	& 85.6	& \textbf{86.8}	& 95.4	& 92.7	& 96.3	& \textbf{97.1}	& 73.0	& 83.1	& \textbf{87.5}\\
Indonesian	& 19\%	& 93.0	& 91.5	& \textbf{93.5}	& 93.4	& 86.2	& \textbf{91.1}	& 91.0	& -		& -		& -		& -				& -		& -		& -\\
Chinese		& 70\%	& 92.5	& 83.5	& 92.8	& \textbf{93.0}	& 88.2	& \textbf{89.2}	& \textbf{89.2}	& 88.2 	& 79.0	& 87.9	& \textbf{88.9}	& \textbf{89.1}	& 87.3	& \textbf{89.1}\\
Persian		& 2\%	& 96.7	& 95.7	& 97.0	& \textbf{97.2}	& 70.1	& 79.8	& \textbf{82.2}	& 96.2	& 91.8	& 96.7	& \textbf{96.9}	& 61.4	& 78.4	& \textbf{83.6}\\
Hebrew		& 9\%	& 95.7	& 95.7	& 96.0	& \textbf{96.2}	& 86.9	& 90.5	& \textbf{91.8}	& 95.0	& 90.3	& 95.1	& \textbf{95.7}	& 88.2	& 92.2	& \textbf{94.1}\\
Romanian	& 27\%	& 96.6	& 95.6	& 97.2	& \textbf{97.3}	& 85.9	& 90.8	& \textbf{91.4}	& 97.5	& 94.2	& 97.9	& \textbf{98.2}	& 90.8	& 94.5	& \textbf{95.6}\\
English		& 4\%	& 94.4	& 93.2	& 94.8	& \textbf{95.3}	& 72.9	& 79.9	& \textbf{84.3}	& 96.3	& 94.7	& 96.5	& \textbf{96.9}	& 82.4	& 87.4	& \textbf{92.7}\\
Arabic		& 27\%	& 96.1	& 95.0	& 96.3	& \textbf{96.6}	& 84.2	& 89.1	& \textbf{90.5}	& 96.1	& 94.2	& 96.1	& \textbf{96.9}	& 79.1	& 84.5	& \textbf{91.4}\\
Hindi		& 4\%	& 96.3	& 93.9	& 96.6	& \textbf{96.8}	& 82.4	& 88.8	& \textbf{90.0}	& 96.6	& 95.1	& 96.8	& \textbf{97.1}	& 89.3	& 94.1	& \textbf{94.3}\\
Italian		& 7\%	& 97.5	& 96.4	& 97.7	& \textbf{97.9}	& 92.8	& 95.9	& \textbf{96.2}	& 98.2	& 96.4	& 98.4	& \textbf{98.7}	& 92.8	& 96.5	& \textbf{97.9}\\
Spanish		& 5\%	& 94.7	& 95.9	& 94.9	& \textbf{95.3}	& 77.0	& 80.9	& \textbf{82.8}	& 97.2	& 95.4	& 97.1	& \textbf{98.0}	& 71.8	& 78.2	& \textbf{81.7}\\
Czech		& 11\%	& 98.4	& 96.6	& 98.5	& \textbf{98.7}	& 93.9	& 95.2	& \textbf{96.0}	& 97.1	& 90.5	& 97.3	& \textbf{97.8}	& 91.9	& 94.0	& \textbf{95.0}\\
\midrule
\textbf{Average} & 16\% & 93.7 & 91.7 & 94.5 &	\textbf{95.1} & 79.8 & 86.7 &\textbf{88.8} & 92.7 &	85.7 & 93.2 &	\textbf{94.1} & 79.4 &	85.4 &	\textbf{89.1}\\
\bottomrule
\end{tabular}
\vspace{6pt}
\caption{Accuracy on part-of-speech tagging (POS) and Micro F1-Score on morphosyntactic attributes (MORPH) on all words (All) and on the OOV words only (OOV) on the Universal Dependencies Dataset 1.4.
In all cases, the tagging net is \citeauthor{plank2016multilingual}'s model (without the special loss, which can explain discrepancies from the original papers).
We compare against randomly assigned embeddings (Rand.), \citeauthor{Zhao2018GeneralizingWE}'s BoSS model and \citeauthor{pinter2017mimicking}'s Mimick (Mim.).
\citeauthor{Zhao2018GeneralizingWE}'s results are those shown in his paper.
Best in each category is highlighted in bold.}
\label{tab:quantitative_results}
\end{table*}

\subsection{Neural Sequence Tagging}

We test our OOV handling layer on the task of assigning part-of-speech tags and morphosyntactic attributes to every word in a sentence.
We insert our layer in the state-of-the-art model proposed by \citeauthor{plank2016multilingual}, which is the same as in \citep{pinter2017mimicking}.
The procedure is outlined in Figure~\ref{fig:task_net}, where the OOV handling layer can either be our model or any other.
Each word is represented by its pre-trained embeddings (purple boxes (w)) from Polyglot or a random embedding (red box (o)).
Our sentence-dependent model handles OOV (``enron'' in the figure) by outputting a predicted substitute embedding (green box (p)). All other words keep their corresponding Polyglot embedding.
The model of \citeauthor{plank2016multilingual} is used (without the correction to the loss to properly evaluate the impact of the OOV handling method) to assign a tag to each word in the sentence.
We used 23 languages of the the Universal Dependencies dataset (1.4) \cite{de2014universal} as a benchmark to evaluate the different configurations.

\vphantom{H\\h\\h}

\colorlet{colorWordEmbeds}{Purple}
\colorlet{colorCharEmbeds}{RedPink}
\colorlet{colorPredEmbed}{PaleGreen}
\colorlet{colorOOV}{RedPink}
\colorlet{colorSoftmax}{gray}
\colorlet{colorTag}{Orange}
\begin{figure}[h!]
\centering
\begin{tikzpicture}[mytip/.tip={>[length=4, width=5.5]},
				    scale=1]

\def\embedWidth{.3}
\def\layerHeight{.42}
\def\layerWidth{4.9}

\def\spaceBetweenWords{.8}
\def\spaceBetweenLayers{.3}
\def\spaceBetweenTextAndEmbed{.2}

\def\sentence{and, when, enron, collapse, the}
\def\nWords{5}
\def\posOOV{3}

\foreach \word [count=\i] in \sentence{
	\pgfmathsetmacro\pos{(\i-(\nWords+1)/2)*(\embedWidth+\spaceBetweenWords)}
	\node[anchor=mid, inner sep=0pt](w \i) at (\pos,0) {\footnotesize \word};
	\ifthenelse{\i=\posOOV}{
		\node[embed=colorOOV](w embed \i) at (\pos,-\spaceBetweenTextAndEmbed) {\footnotesize o};
	}{
		\node[embed=colorWordEmbeds](w embed \i) at (\pos,-\spaceBetweenTextAndEmbed) {\footnotesize w};
	}
}

\node[draw=black, thick, minimum width=\layerWidth cm, minimum height=\layerHeight mm, anchor=north](OOV handling) at ([yshift=-\spaceBetweenLayers cm]w embed 1.south -| 0,0) {OOV handling layer};
\foreach \word [count=\i] in \sentence{
	\draw[-mytip] (w embed \i) -- (w embed \i |- OOV handling.north);
}

\foreach \word [count=\i] in \sentence{
	\ifthenelse{\i=\posOOV}{
		\node[embed=colorPredEmbed](p embed \i) at ([yshift=-\spaceBetweenLayers cm]w embed \i |- OOV handling.south) {\footnotesize p};
	}{
		\node[embed=colorWordEmbeds](p embed \i) at ([yshift=-\spaceBetweenLayers cm]w embed \i |- OOV handling.south) {\footnotesize w};
	}
	\draw[-mytip] (w embed \i |- OOV handling.south) -- (p embed \i);
}

\node[draw=black, thick, minimum width=\layerWidth cm, minimum height=\layerHeight mm, anchor=north](tagging net) at ([yshift=-\spaceBetweenLayers cm]p embed 1.south -| 0,0) {Tagging neural net};
\foreach \word [count=\i] in \sentence{
	\draw[-mytip] (p embed \i) -- (p embed \i |- tagging net.north);
}

\foreach \word [count=\i] in \sentence{
	\node[embed=colorTag](t embed \i) at ([yshift=-\spaceBetweenLayers cm]p embed \i |- tagging net.south) {\footnotesize t};
	\draw[-mytip] (p embed \i |- tagging net.south) -- (t embed \i);
}

\end{tikzpicture}
\caption{The sequence tagging neural net and where our model stands.}\label{fig:task_net}
\end{figure}
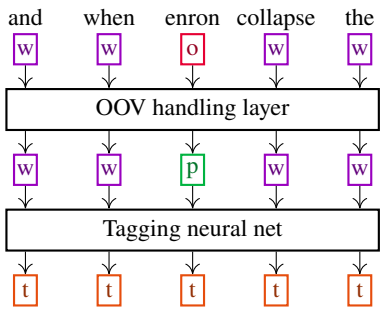

\subsection{Training}

A consequence of the proposed approach is that our model removes the need for a separate pre-training step as in 
previous work \citep{pinter2017mimicking, Zhao2018GeneralizingWE, Schick2018LearningSR},
since it learns
how to generate context-dependent and task-specific representations using the sequence tagging task.
Thus, we are not using the set of pre-trained embeddings as a signal at all as proposed in previous architectures making the overall training procedure simpler.
The only signal our model receives is from the downstream tags.

Because the model can only learn from OOV words, it may be possible that the number of OOV words in the training set is not sufficient to train properly.
To solve this problem, we sample $k$ words from the vocabulary $\mathcal{V}$ at each mini-batch during training and let our architecture predict their embeddings.
This way our model receives a decent amount of gradient needed to update properly the parameters.
Moreover, this approach can be seen as \textit{word dropout} \cite{goldberg2017neural_chap8}, making the network more resilient to noisy data.
In our experiments, we set $k$ to be 15\% of $|\mathcal{V}|$.

We used Adam \cite{Kingma2015AdamAM} as the optimizer, a mini-batch size of 32 examples and a normal Kaiming distribution \cite{He2015DelvingDI} to initialize properly the network's parameters.


\section{Results and discussion}
\label{sec:results}

We experiment on the same subset of 23 languages of the Universal Dependencies corpus as \newcite{pinter2017mimicking} for a fair comparison.
We compare our model with three other OOV handling methods: random embeddings assignment (which is the vanilla model proposed by \citeauthor{plank2016multilingual}), BoSS and Mimick embeddings, which held the state of the art on this dataset before us.
It would be unfair to compare against \citet{Schick2018LearningSR} since our setting does not provide enough training data.
The results are presented in Table~\ref{tab:quantitative_results}, where we considered performance overall and over the OOV words that are in the test set (but not in the training set), which really shows the relevance of our model.


Our model performs better than the other OOV handling methods on all languages but one on both tasks.
It has an average relative gain over Mimick's performances of 0.66\% on the POS task with highest gains on Kazakh (3.64\%) and Latvian (1.64\%).
The average relative gain on the MORPH task is 1.02\% with the highest on Latvian (3.1\%) and Kazakh (2.5\%).
The relative gains are even more significant when looking only at the OOV words POS accuracy (2.54\%) and MORPH F1-Score (4.84\%).
The highest gains on OOV POS accuracy are Latvian (11.53\%) and Basque (5.8\%).
On the OOV MORPH F1-Score our model has impressive gains on Latvian (17.86\%) and Kazakh (14.29\%). 

The fact that our model helps significantly in languages where the OOV rate is high, such as Basque and Arabic, shows that our model really provides useful word representations for OOV words.
Moreover, it is notable that our additional layer allows a state-of-the-art model to gain in performance even in languages where the OOV rate is very low, like Persian and English, showing once more the importance of handling properly OOV words in NLP tasks.
Since our model uses attention over the context and the characters of a particular OOV, we provide a visualization of the mechanism in the Appendix~\ref{app:attention}.
While there are morphological and contextual patterns emanating from the visualization, their interpretation is unclear.

\section{Conclusion}

In this paper, we showed that leveraging the characters of an unknown word and the context in which it appears help to generate useful sentence-dependent and task-specific representations for NLP tagging tasks.
Our layer with a simple attention mechanism improved the performances of the model on essentially every languages tested.
Moreover, our model can be seen as a drop-in embedding layer without the need for pre-training on a specific corpus and we seek to apply it in other tasks in the future.
Also, since its number of parameters does not grow with the vocabulary size, we hope to use our model as a compression factor over the embedding table, similar to \citet{Ravi2018SelfGoverningNN}.



\clearpage
\bibliographystyle{acl_natbib}
\bibliography{acl2019}

\clearpage
\appendix

\onecolumn

\section{Visualization of the attention mechanism}\label{app:attention}

To peek into the model internals, we inspect the weights from the attention mechanism which weighs the hidden state associated each word of the context and each character of the OOV word.
We observed much difference in the behavior of the attention mechanism, which are generally consistent within each language.
Examples of some of these are shown in Table~\ref{tab:qualitative_results}\footnote{All examples can be found at \\ \url{https://goo.gl/CQuX16}.}.
We note that this visualization has been produced on the joint POS and MORPH tagging task, and could thus hinder the interpretability in some cases.

Some observed behaviors on the context include: focus spread variably on most of the words in the sentence (English, Italian, Greek, Russian), focus on words in different position in the sentence (Hebrew, Spanish, Latvian, Turkish), focus on only a few words of the sentence (Arabic, Chinese).
Similar phenomena are witnessed for the characters composing the OOV word depending on the language.

Amongst the examples, the Latvian (LV) example presented in Table~\ref{tab:qualitative_results} is interesting morphologically.
Latvian has several declensions related to POS and to MORPH attributes, so it is natural that the attention is aimed at the end of the OOV word, explaining why our model helps so much in tagging Latvian.
A similar observation can be made for Russian (RU), where the suffix \foreignlanguage{russian}{-ого} is important for determining the genitive case (MORPH) of this adjective (POS) and it is well captured by the model, getting in total 30\% of the attention.

However, while some attention patterns can be interpreted, not all behaviors are easily understood.
Take for example Indonesian (ID).
The capital letters at the beginning of the words are well captured by the attention, however the context seems uninformative, where the attention is spread evenly on all the words of the sentence.
This may explain why the model performs comparably to other methods on this language specifically.

\begin{table*}[h]
\vspace{6pt}
\centering
\small
\begin{tabular}{p{15.5cm}}
\toprule

\textbf{KK}: \foreignlanguage{russian}{
\att{Жанына}{69.98000144958496} \att{мылтық}{29.980000853538513} 
\textbf{[
\att{б}{3.0899999663233757}%
\att{а}{3.1300000846385956}%
\att{т}{7.240000367164612}%
\att{қ}{33.36000144481659}%
\att{а}{25.42000114917755}%
\att{н}{23.399999737739563}%
\att{д}{2.710000053048134}%
\att{а}{1.6599999740719795}%
]}
\att{,}{.039999998989515007}
\att{анасы}{0}
\att{қызын}{0}
\att{тастай}{0}
\att{беріп}{0}
\att{,}{0}
\att{алыса}{0}
\att{кетті}{0}
\att{.}{0}
}
\\[5pt]

\textbf{LV}: {\transparent{0.0031999999191612005}\colorbox{red}{\transparent{1.0}{{\strut Ir}}}} {\transparent{0.006099999882280827}\colorbox{red}{\transparent{1.0}{{\strut jāskatās}}}} {\transparent{0.00009}\colorbox{red}{\transparent{1.0}{{\strut ,}}}} {\transparent{0.0}\colorbox{red}{\transparent{1.0}{{\strut lai}}}} {\transparent{0.00019999999494757503}\colorbox{red}{\transparent{1.0}{{\strut pakalpojumi}}}} {\transparent{0.0010000000474974513}\colorbox{red}{\transparent{1.0}{{\strut būtu}}}}
\textbf{[{\transparent{0.0}\colorbox{red}{\transparent{1.0}{{\strut k}}}}{\transparent{0.0}\colorbox{red}{\transparent{1.0}{{\strut o}}}}{\transparent{0.0}\colorbox{red}{\transparent{1.0}{{\strut n}}}}{\transparent{0.0}\colorbox{red}{\transparent{1.0}{{\strut k}}}}{\transparent{0.0}\colorbox{red}{\transparent{1.0}{{\strut u}}}}{\transparent{0.0}\colorbox{red}{\transparent{1.0}{{\strut r}}}}{\transparent{0.0}\colorbox{red}{\transparent{1.0}{{\strut ē}}}}{\transparent{0.0}\colorbox{red}{\transparent{1.0}{{\strut t}}}}{\transparent{0.00009}\colorbox{red}{\transparent{1.0}{{\strut s}}}}{\transparent{0.0010000000474974513}\colorbox{red}{\transparent{1.0}{{\strut p}}}}{\transparent{0.00139999995008111}\colorbox{red}{\transparent{1.0}{{\strut ē}}}}{\transparent{0.010700000450015068}\colorbox{red}{\transparent{1.0}{{\strut j}}}}{\transparent{0.25589999556541443}\colorbox{red}{\transparent{1.0}{{\strut ī}}}}{\transparent{0.7023000121116638}\colorbox{red}{\transparent{1.0}{{\strut g}}}}{\transparent{0.02850000001490116}\colorbox{red}{\transparent{1.0}{{\strut i}}}}]}
{\transparent{0.3928000032901764}\colorbox{red}{\transparent{1.0}{{\strut cenas}}}} {\transparent{0.16670000553131104}\colorbox{red}{\transparent{1.0}{{\strut ziņā}}}} {\transparent{0.42989999055862427}\colorbox{red}{\transparent{1.0}{{\strut .}}}}
\\[5pt]

\textbf{TR}: {\transparent{0.018300000578165054}\colorbox{red}{\transparent{1.0}{{\strut Bu}}}} {\transparent{0.8120999932289124}\colorbox{red}{\transparent{1.0}{{\strut kamp}}}} {\transparent{0.13670000433921814}\colorbox{red}{\transparent{1.0}{{\strut alanında}}}}
\textbf{[{\transparent{0.0}\colorbox{red}{\transparent{1.0}{{\strut d}}}}{\transparent{0.00009}\colorbox{red}{\transparent{1.0}{{\strut e}}}}{\transparent{0.14339999854564667}\colorbox{red}{\transparent{1.0}{{\strut p}}}}{\transparent{0.5260000228881836}\colorbox{red}{\transparent{1.0}{{\strut o}}}}{\transparent{0.1826999932527542}\colorbox{red}{\transparent{1.0}{{\strut l}}}}{\transparent{0.13580000400543213}\colorbox{red}{\transparent{1.0}{{\strut a}}}}{\transparent{0.0}\colorbox{red}{\transparent{1.0}{{\strut r}}}}{\transparent{0.012000000104308128}\colorbox{red}{\transparent{1.0}{{\strut ı}}}}{\transparent{0.0}\colorbox{red}{\transparent{1.0}{{\strut n}}}}]}
{\transparent{0.011300000362098217}\colorbox{red}{\transparent{1.0}{{\strut dışında}}}} {\transparent{0.007899999618530273}\colorbox{red}{\transparent{1.0}{{\strut iki}}}} {\transparent{0.007499999832361937}\colorbox{red}{\transparent{1.0}{{\strut ev}}}} {\transparent{0.006200000178068876}\colorbox{red}{\transparent{1.0}{{\strut var}}}} {\transparent{0.0}\colorbox{red}{\transparent{1.0}{{\strut .}}}}\\[5pt]

\textbf{EL}: \foreignlanguage{greek}{
{\transparent{0.13079999387264252}\colorbox{red}{\transparent{1.0}{{\strut Αφήστε}}}} {\transparent{0.14000000059604645}\colorbox{red}{\transparent{1.0}{{\strut το}}}} {\transparent{0.11829999834299088}\colorbox{red}{\transparent{1.0}{{\strut ένα}}}} {\transparent{0.12039999663829803}\colorbox{red}{\transparent{1.0}{{\strut τρίτο}}}} {\transparent{0.10809999704360962}\colorbox{red}{\transparent{1.0}{{\strut του}}}} {\transparent{0.125}\colorbox{red}{\transparent{1.0}{{\strut χρόνου}}}}
\textbf{[{\transparent{0.21250000596046448}\colorbox{red}{\transparent{1.0}{{\strut α}}}}{\transparent{0.12139999866485596}\colorbox{red}{\transparent{1.0}{{\strut γ}}}}{\transparent{0.21070000529289246}\colorbox{red}{\transparent{1.0}{{\strut ό}}}}{\transparent{0.033900000154972076}\colorbox{red}{\transparent{1.0}{{\strut ρ}}}}{\transparent{0.07029999792575836}\colorbox{red}{\transparent{1.0}{{\strut ε}}}}{\transparent{0.03779999911785126}\colorbox{red}{\transparent{1.0}{{\strut υ}}}}{\transparent{0.17309999465942383}\colorbox{red}{\transparent{1.0}{{\strut σ}}}}{\transparent{0.052299998700618744}\colorbox{red}{\transparent{1.0}{{\strut η}}}}{\transparent{0.08799999952316284}\colorbox{red}{\transparent{1.0}{{\strut ς}}}}]}
{\transparent{0.07360000163316727}\colorbox{red}{\transparent{1.0}{{\strut για}}}} {\transparent{0.0786999985575676}\colorbox{red}{\transparent{1.0}{{\strut πραγματικές}}}} {\transparent{0.04129999876022339}\colorbox{red}{\transparent{1.0}{{\strut συζητήσεις}}}}
{\transparent{0.06379999965429306}\colorbox{red}{\transparent{1.0}{{\strut .}}}}
}\\[5pt]

\textbf{RU}: \foreignlanguage{russian}{
{\transparent{0.1340000033378601}\colorbox{red}{\transparent{1.0}{{\strut Недалеко}}}} {\transparent{0.1469999998807907}\colorbox{red}{\transparent{1.0}{{\strut от}}}}
\textbf{[{\transparent{0.020400000736117363}\colorbox{red}{\transparent{1.0}{{\strut д}}}}{\transparent{0.13259999454021454}\colorbox{red}{\transparent{1.0}{{\strut а}}}}{\transparent{0.050599999725818634}\colorbox{red}{\transparent{1.0}{{\strut о}}}}{\transparent{0.04960000142455101}\colorbox{red}{\transparent{1.0}{{\strut с}}}}{\transparent{0.1518000066280365}\colorbox{red}{\transparent{1.0}{{\strut к}}}}{\transparent{0.2207999974489212}\colorbox{red}{\transparent{1.0}{{\strut о}}}}{\transparent{0.07819999754428864}\colorbox{red}{\transparent{1.0}{{\strut г}}}}{\transparent{0.29589998722076416}\colorbox{red}{\transparent{1.0}{{\strut о}}}}]}
{\transparent{0.14880000054836273}\colorbox{red}{\transparent{1.0}{{\strut храма}}}} {\transparent{0.05689999833703041}\colorbox{red}{\transparent{1.0}{{\strut Чэнхуанмяо}}}} {\transparent{0.1395999938249588}\colorbox{red}{\transparent{1.0}{{\strut раньше}}}} {\transparent{0.08919999748468399}\colorbox{red}{\transparent{1.0}{{\strut находился}}}} {\transparent{0.13770000636577606}\colorbox{red}{\transparent{1.0}{{\strut рынок}}}} {\transparent{0.1467999964952469}\colorbox{red}{\transparent{1.0}{{\strut .}}}}
}



\\
\textbf{ID:}
\textbf{[{\transparent{0.33149999380111694}\colorbox{red}{\transparent{1.0}{{\strut S}}}}{\transparent{0.017500000074505806}\colorbox{red}{\transparent{1.0}{{\strut i}}}}{\transparent{0.02850000001490116}\colorbox{red}{\transparent{1.0}{{\strut n}}}}{\transparent{0.010400000028312206}\colorbox{red}{\transparent{1.0}{{\strut d}}}}{\transparent{0.04129999876022339}\colorbox{red}{\transparent{1.0}{{\strut a}}}}{\transparent{0.04529999941587448}\colorbox{red}{\transparent{1.0}{{\strut n}}}}{\transparent{0.0738999992609024}\colorbox{red}{\transparent{1.0}{{\strut g}}}}{\transparent{0.03180000185966492}\colorbox{red}{\transparent{1.0}{{\strut h}}}}{\transparent{0.2020999938249588}\colorbox{red}{\transparent{1.0}{{\strut a}}}}{\transparent{0.149399995803833}\colorbox{red}{\transparent{1.0}{{\strut y}}}}{\transparent{0.06830000132322311}\colorbox{red}{\transparent{1.0}{{\strut u}}}}]}
{\transparent{0.0706000030040741}\colorbox{red}{\transparent{1.0}{{\strut adalah}}}} {\transparent{0.0722000002861023}\colorbox{red}{\transparent{1.0}{{\strut desa}}}} {\transparent{0.07729999721050262}\colorbox{red}{\transparent{1.0}{{\strut di}}}} {\transparent{0.0575999990105629}\colorbox{red}{\transparent{1.0}{{\strut kecamatan}}}} {\transparent{0.06499999761581421}\colorbox{red}{\transparent{1.0}{{\strut Takokak}}}} {\transparent{0.08049999922513962}\colorbox{red}{\transparent{1.0}{{\strut Cianjur}}}} {\transparent{0.08100000023841858}\colorbox{red}{\transparent{1.0}{{\strut Jawa}}}} {\transparent{0.0778999999165535}\colorbox{red}{\transparent{1.0}{{\strut Barat}}}} {\transparent{0.07689999788999557}\colorbox{red}{\transparent{1.0}{{\strut Indonesia}}}} {\transparent{0.09709999710321426}\colorbox{red}{\transparent{1.0}{{\strut .}}}}
\\[5pt]

\textbf{EN}: {\transparent{0.0771000012755394}\colorbox{red}{\transparent{1.0}{{\strut the}}}} {\transparent{0.12229999899864197}\colorbox{red}{\transparent{1.0}{{\strut economy}}}} {\transparent{0.04879999905824661}\colorbox{red}{\transparent{1.0}{{\strut is}}}} {\transparent{0.02070000022649765}\colorbox{red}{\transparent{1.0}{{\strut down}}}} {\transparent{0.07639999687671661}\colorbox{red}{\transparent{1.0}{{\strut and}}}} {\transparent{0.03750000149011612}\colorbox{red}{\transparent{1.0}{{\strut when}}}}
\textbf{[{\transparent{0.0}\colorbox{red}{\transparent{1.0}{{\strut e}}}}{\transparent{0.009800000116229057}\colorbox{red}{\transparent{1.0}{{\strut n}}}}{\transparent{0.9901999831199646}\colorbox{red}{\transparent{1.0}{{\strut r}}}}{\transparent{0.0}\colorbox{red}{\transparent{1.0}{{\strut o}}}}{\transparent{0.0}\colorbox{red}{\transparent{1.0}{{\strut n}}}}]}
{\transparent{0.09709999710321426}\colorbox{red}{\transparent{1.0}{{\strut collapses}}}} {\transparent{0.0820000022649765}\colorbox{red}{\transparent{1.0}{{\strut ,}}}} {\transparent{0.07609999924898148}\colorbox{red}{\transparent{1.0}{{\strut the}}}} {\transparent{0.07079999893903732}\colorbox{red}{\transparent{1.0}{{\strut energy}}}} {\transparent{0.060499999672174454}\colorbox{red}{\transparent{1.0}{{\strut industry}}}} {\transparent{0.03579999879002571}\colorbox{red}{\transparent{1.0}{{\strut is}}}} {\transparent{0.02280000038444996}\colorbox{red}{\transparent{1.0}{{\strut going}}}} {\transparent{0.0729999989271164}\colorbox{red}{\transparent{1.0}{{\strut to}}}} {\transparent{0.024399999529123306}\colorbox{red}{\transparent{1.0}{{\strut be}}}} {\transparent{0.03880000114440918}\colorbox{red}{\transparent{1.0}{{\strut in}}}} {\transparent{0.019500000402331352}\colorbox{red}{\transparent{1.0}{{\strut a}}}} {\transparent{0.007699999958276749}\colorbox{red}{\transparent{1.0}{{\strut world}}}} {\transparent{0.008500000461935997}\colorbox{red}{\transparent{1.0}{{\strut of}}}} {\transparent{0.00009}\colorbox{red}{\transparent{1.0}{{\strut hurt}}}} {\transparent{0.0}\colorbox{red}{\transparent{1.0}{{\strut .}}}}
\\[5pt]

\textbf{ES}: {\transparent{0.1428000032901764}\colorbox{red}{\transparent{1.0}{{\strut Por}}}} {\transparent{0.17720000445842743}\colorbox{red}{\transparent{1.0}{{\strut eso}}}} {\transparent{0.07479999959468842}\colorbox{red}{\transparent{1.0}{{\strut ,}}}} {\transparent{0.16500000655651093}\colorbox{red}{\transparent{1.0}{{\strut esto}}}} {\transparent{0.005900000222027302}\colorbox{red}{\transparent{1.0}{{\strut se}}}}
\textbf{[{\transparent{0.10279999673366547}\colorbox{red}{\transparent{1.0}{{\strut p}}}}{\transparent{0.003800000064074993}\colorbox{red}{\transparent{1.0}{{\strut r}}}}{\transparent{0.148499995470047}\colorbox{red}{\transparent{1.0}{{\strut o}}}}{\transparent{0.02710000053048134}\colorbox{red}{\transparent{1.0}{{\strut f}}}}{\transparent{0.066600002348423}\colorbox{red}{\transparent{1.0}{{\strut u}}}}{\transparent{0.05570000037550926}\colorbox{red}{\transparent{1.0}{{\strut n}}}}{\transparent{0.23479999601840973}\colorbox{red}{\transparent{1.0}{{\strut d}}}}{\transparent{0.16670000553131104}\colorbox{red}{\transparent{1.0}{{\strut i}}}}{\transparent{0.006899999920278788}\colorbox{red}{\transparent{1.0}{{\strut z}}}}{\transparent{0.07370000332593918}\colorbox{red}{\transparent{1.0}{{\strut a}}}}{\transparent{0.028300000354647636}\colorbox{red}{\transparent{1.0}{{\strut r}}}}{\transparent{0.0851999968290329}\colorbox{red}{\transparent{1.0}{{\strut a}}}}]}
{\transparent{0.00009}\colorbox{red}{\transparent{1.0}{{\strut con}}}} {\transparent{0.0010999999940395355}\colorbox{red}{\transparent{1.0}{{\strut la}}}} {\transparent{0.17749999463558197}\colorbox{red}{\transparent{1.0}{{\strut ayuda}}}} {\transparent{0.11349999904632568}\colorbox{red}{\transparent{1.0}{{\strut inestimable}}}} {\transparent{0.00019999999494757503}\colorbox{red}{\transparent{1.0}{{\strut de}}}} {\transparent{0.03909999877214432}\colorbox{red}{\transparent{1.0}{{\strut todos}}}} {\transparent{0.10279999673366547}\colorbox{red}{\transparent{1.0}{{\strut nosotros}}}} {\transparent{0.0}\colorbox{red}{\transparent{1.0}{{\strut .}}}}
\\
\bottomrule
\end{tabular}
\vspace{6pt}
\caption{The behavior of the attention mechanisms (characters and contexts) trained on both tasks differs between languages. OOV words are in bold between brackets. All shown examples are well classified by our model. A slight temperature has been added to the softmax to help distinguish the relative importance of each element. Shown languages are Kazakh (KK), Latvian (LV), Turkish (TR), Greek (EL), Russian (RU), Indonesian (ID), English (EN), Spanish (ES).}
\label{tab:qualitative_results}
\end{table*}

\end{document}